\documentclass[letterpaper, 10 pt, journal, twoside]{IEEEtran}  

\IEEEoverridecommandlockouts                              

\usepackage[hidelinks,bookmarks=false]{hyperref}

\usepackage{amsmath}
\usepackage{amssymb}
\usepackage{algorithm}
\usepackage{algpseudocode}
\usepackage{graphicx}
\usepackage{multirow} 
\usepackage{nicematrix}
\newcolumntype{C}{>{\centering\arraybackslash}X}
\usepackage{acronym}
\usepackage{cleveref}
\usepackage{subcaption}
\usepackage{siunitx}
\usepackage{svg}
\usepackage{url}
\usepackage{cite}
\usepackage{pgf} 
\usepackage{algorithm}
\usepackage{algpseudocode}
\usepackage{bm}       
\usepackage{adjustbox}

\usepackage{graphicx}
\usepackage{overpic}
\usepackage{xcolor}
\usepackage{listings}
\usepackage{cleveref}

\usepackage{tikz}              
\usetikzlibrary{positioning, arrows.meta, backgrounds}

\crefname{figure}{Fig.}{Figs.}
\Crefname{figure}{Fig.}{Figs.}
\crefname{equation}{Eq.}{Eqs.}
\Crefname{equation}{Eq.}{Eqs.}

\usepackage[utf8]{inputenc}  
\DeclareUnicodeCharacter{2212}{-}

\acrodef{pom}[POM]{Polyoxymethylene}
\acrodef{pa12}[PA12]{Nylon-12}
\acrodef{sls}[SLS]{Selective Laser Sintering}

\acrodef{iekf}[IEKF]{Invariant Extended Kalman Filter}

\acrodef{bim}[BIM]{Building Information Modeling}
\acrodef{vjm}[VJM]{Virtual Joint Method}

\acrodef{loocv}[LOOCV]{Leave-one-out cross-validation}

\DeclareMathOperator*{\argmin}{argmin}

\newcommand{\T}[2]{\mathbf{T}_{\!\mathcal{#1},\mathcal{#2}}}
\newcommand{\Traw}[2]{\mathbf{T}_{\!#1,#2}}
\newcommand{\p}[2]{\mathbf{p}_{\mathcal{#1},\mathcal{#2}}}
\newcommand{\praw}[2]{\mathbf{p}_{#1,#2}}
\newcommand{\R}[2]{\mathbf{R}_{\!\mathcal{#1},\mathcal{#2}}}

\newcommand{\extractpart}[2]{\left[ #1\right]_{#2}}
\newcommand{\pospart}[1]{\extractpart{#1}{\mathbf{p}}}
\newcommand{\rotpart}[1]{\extractpart{#1}{\mathbf{R}}}

\usepackage{soul,xcolor}
\usepackage{xparse}
\sethlcolor{red!50} 
\usepackage{etoolbox} 

\newcommand{\threeSPR}{\mbox{3-S\underline{P}R}}
\newcommand{\threeRPS}{\mbox{3-R\underline{P}S}}

\newcommand{\Rthree}{\mathbb{R}^{3}}

\newcommand{\SOthree}{\text{SO}(3)}
\newcommand{\SEthree}{\text{SE}(3)}

\title{Tripody: An Overconstrained \threeSPR{}-like Parallel Robot for High‑Reach Construction Tasks}

\author{Julien Kindle$^{1,2}$, Jakub Raczy$^{2}$, Riccardo Balbi$^{2}$, Andrea Alessandretti$^{2}$, Cesar Cadena$^{1}$ and Marco Hutter$^{1}$
\thanks{Manuscript received: March 13, 2026; Revised May 28, 2026; Accepted July 2, 2026.}
\thanks{This paper was recommended for publication by Editor Cosimo Della Santina upon evaluation of the Associate Editor and Reviewers’ comments.}
\thanks{This work was supported by Hilti AG, Schaan, Liechtenstein.}%
\thanks{$^{1}$Affiliated with the Robotic Systems Lab, ETH Zurich, Switzerland}%
\thanks{$^{2}$Affiliated with Hilti AG, Schaan, Liechtenstein}%
\thanks{Correspondence: {\tt\small jkindle@ethz.ch}}%
\thanks{Digital Object Identifier (DOI): see top of this page.}
}

\begin{document}

\begin{titlepage}
  \onecolumn
  \vspace*{2cm}

  \begin{center}

    {\Large \bfseries Preprint Version}\\[1em]

    {\bfseries
    Tripody: An Overconstrained \threeSPR{}-like Parallel Robot \\
    for High-Reach Construction Tasks}\\[1em]

    \textit{Accepted in IEEE Robotics and Automation Letters on July 2, 2026}

    \vspace{2em}

    \begin{minipage}{0.95\textwidth}
      \small
      \textcopyright 2026 IEEE. Personal use of this material is
      permitted. Permission from IEEE must be obtained for all other
      uses, in any current or future media, including
      reprinting/republishing this material for advertising or
      promotional purposes, creating new collective works, for resale
      or redistribution to servers or lists, or reuse of any
      copyrighted component of this work in other works.\\
      \textbf{DOI:}
      \href{https://doi.org/10.1109/LRA.2026.3713724}
      {10.1109/LRA.2026.3713724}
    \end{minipage}

    \vspace{2em}

    \hrule
    \vspace{1em}

    \textbf{Please cite this publication as:}

    \vspace{0.5em}

    \begin{minipage}{0.95\textwidth}
      \small
      J.~Kindle, J.~Raczy, R.~Balbi, A.~Alessandretti,
      C.~Cadena and M.~Hutter,
      ``\textit{Tripody: An Overconstrained
      \threeSPR{}-like Parallel Robot for High-Reach
      Construction Tasks},''
      in IEEE Robotics and Automation Letters,
      vol.~11, no.~9, pp.~10497-10504, Jul.~2026,
      doi:~10.1109/LRA.2026.3713724.
    \end{minipage}

    \vspace{1em}
    \hrule

    \vspace{2em}

    \textbf{Bib\TeX~entry:}\\[0.5em]

\begin{lstlisting}
@ARTICLE{kindle_tripody,
  author={Kindle, Julien and Raczy, Jakub and Balbi, Riccardo and Alessandretti, Andrea and Cadena, Cesar and Hutter, Marco},
  journal={IEEE Robotics and Automation Letters},
  title={{Tripody: An Overconstrained 3-S\underline{P}R-like Parallel Robot for High-Reach Construction Tasks}},
  year={2026},
  volume={11},
  number={9},
  pages={10497-10504},
  keywords={Robotics and Automation in Construction;Mechanism Design;Parallel Robots},
  doi={10.1109/LRA.2026.3713724}
}
\end{lstlisting}

  \end{center}

  \vfill
\end{titlepage}

\bstctlcite{BSTcontrol}

\markboth{IEEE Robotics and Automation Letters. Preprint Version. Accepted July, 2026}
{Kindle \MakeLowercase{\textit{et al.}}: Tripody: A Parallel Robot for High-Reach Construction} 

\maketitle

\begin{abstract}
Many ceiling construction tasks still rely on heavy serial manipulators that are difficult to deploy in cluttered interiors, motivating lightweight, field-ready alternatives that reach ceiling height while maintaining millimeter-level accuracy and the stiffness demanded by overhead tool loads. We introduce \emph{Tripody}, a wheeled \mbox{3-DoF} parallel robot for high-reach tasks that replaces the base spherical joints of a classical \threeSPR{} (3~legs; S:~base spherical joint; \underline{P}:~actuated prismatic joint; R:~end-effector revolute joint) morphology with universal joints, intentionally overconstraining the mechanism; small, distributed elastic deflections absorb the resulting incompatibilities, preserving predominantly translational motion. The \SI{33}{\kilo\gram} system extends from \SI{1.7}{\meter} to \SI{3.4}{\meter} in height, supports a continuous \SI{32}{\kilo\gram} payload, and offers a modular end-effector interface for ceiling operations. We detail the mechanical design - including custom linear actuators and a kinematic-compatibility analysis - and a control stack for accurate positioning that combines SE(3) state estimation, forward kinematics, and task-space control. In experiments, Tripody exhibits similar in-plane stiffness to a spherical-base variant but substantially higher torsional stiffness - an increase of \SI{67}{\percent} at \SI{1.7}{\meter}, \SI{196}{\percent} at \SI{2.6}{\meter}, and \SI{454}{\percent} at \SI{3.4}{\meter} - while maintaining negligible cross-axis coupling. Closed-loop positioning with a total station converges below \SI{0.6}{\milli\meter} across the entire workspace; pure model extrapolation achieves a 95th-percentile error of \SI{2.7}{\milli\meter} (max \SI{3.6}{\milli\meter}). Finally, we demonstrate task-level ceiling-drilling feasibility in an open-loop study by drilling a 15-hole pattern with \SI{4.5}{\milli\meter} maximum relative hole-position error after rigid alignment. These results support overconstrained, compliance-absorbing \threeSPR{}-like architectures as a practical path to lightweight, high-reach, millimeter-accurate construction robots.
\end{abstract}


\section{INTRODUCTION}\label{sec:introduction}
\begin{figure}[t]
  \includegraphics[width=\linewidth]{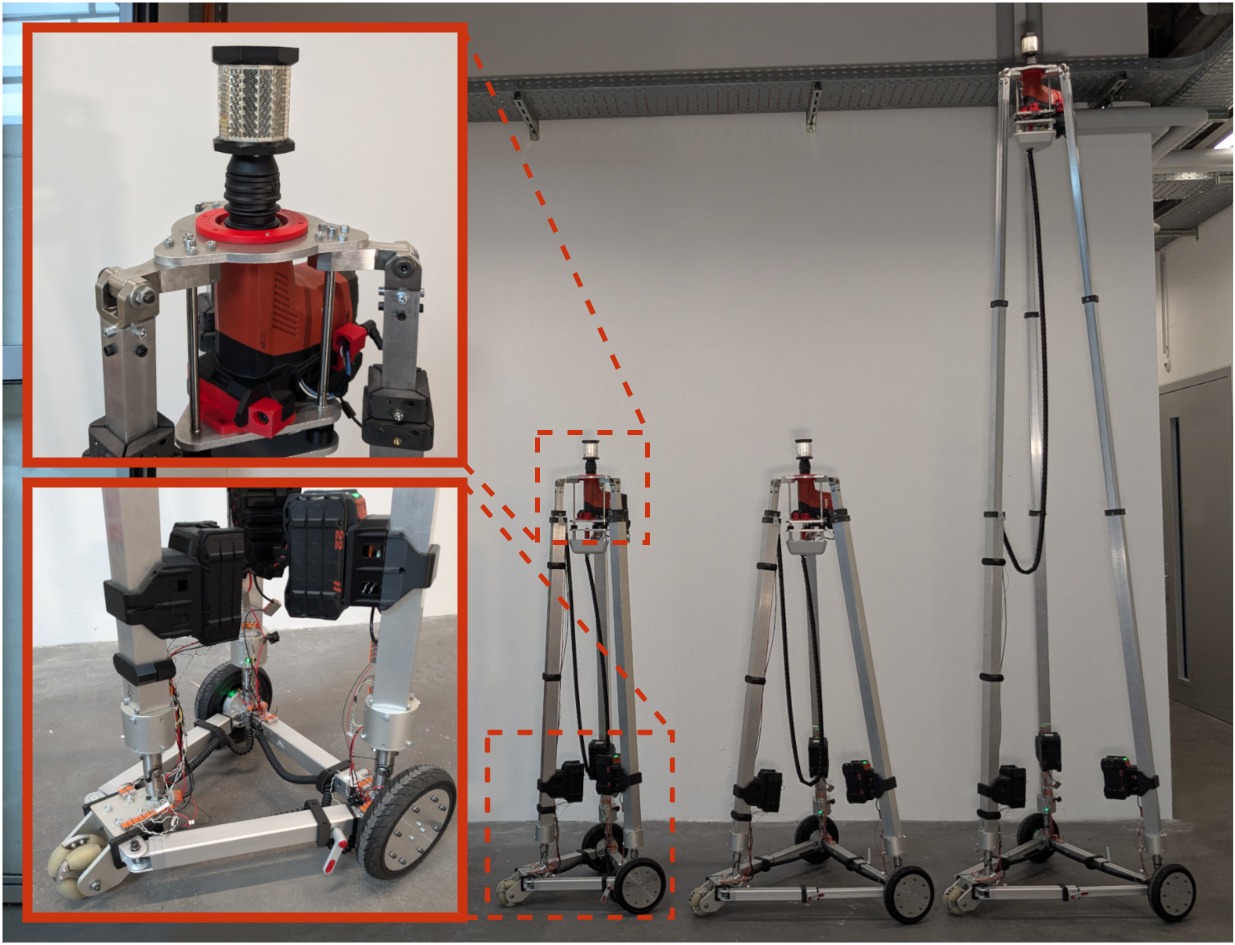}
  \caption{Tripody equipped with a drilling end-effector shown in transportation, retracted, and extended configurations. Insets highlight the base and end-effector assemblies.}
  \label{fig:tripody}
\end{figure}

\IEEEPARstart{M}{ANY} construction tasks involve sustained overhead ceiling work at significant heights, increasing both the immediate risk of injury and the likelihood of long-term musculoskeletal strain. In the United States, an estimated 76\% of construction workers perform work at elevation at least monthly, and 37\% climb ladders or scaffolds for half their work time or more~\cite{cpwr2018chartbook}. Falls remain the leading cause of fatal injuries in the construction industry, accounting for 39\% of worker deaths in 2023, and rank second among all unintentional injury deaths worldwide, claiming roughly 684,000 lives each year~\cite{osha_stop_falls,WHO2021Falls}. Beyond the immediate danger of falls, repeated overhead work exerts substantial ergonomic stress on the neck, shoulders, and upper back. Tasks that require holding the arms above shoulder level - such as ceiling installation or overhead drilling - force workers into awkward, non‐neutral postures that elevate joint loading and compromise circulation~\cite{CHARLES2018125}. When combined with hand‐arm vibration (from drilling or grinding) and exposure to respirable crystalline silica, these factors lead to both acute injuries and chronic musculoskeletal disorders.

Robotic systems have begun to alleviate these hazards by automating high‐elevation tasks. Commercial platforms like the \textit{Hilti Jaibot}~\cite{jaibot_patent} and \textit{fischer Baubot}~\cite{baubot_patent} autonomously mark and drill anchor points, significantly reducing manual overhead drilling. However, these robots typically exceed \SI{500}{\kilo\gram}, demand on‐site cranes 
for deployment, and struggle to access confined or cluttered renovation environments.

This motivates a lightweight platform capable of reaching ceiling heights in confined spaces while maintaining millimeter-level accuracy and the stiffness required for overhead construction tasks. In response, we introduce Tripody~(\Cref{fig:tripody}), a lightweight, wheeled, and field-deployable \mbox{3-DoF} parallel robot with U\underline{P}R legs (U: base universal joint; \underline{P}: actuated prismatic joint; R: end-effector revolute joint) that resembles a classic \threeSPR{} topology but is intentionally overconstrained by replacing the base spherical joints with universal joints. The resulting overconstraint is absorbed by small elastic deflections, preserving predominantly translational end-effector motion while markedly increasing resistance to tool-induced twisting and vibration. Tripody can be carried by two workers, extends from \SI{1.7}{\meter} to \SI{3.4}{\meter} in height, sustains a continuous payload of \SI{32}{\kilo\gram}, and reaches horizontal end-effector speeds up to \SI{1.3}{\meter\per\second}.


The main contributions of this work are:
\begin{itemize}
  \item a lightweight, wheeled, and field-deployable robot for high-reach ceiling tasks that intentionally removes one passive rotational degree of freedom per leg from a \threeSPR{} morphology; the resulting overconstraint is absorbed by structural compliance.
  \item a control stack for high-accuracy positioning, combining SE(3) state estimation, geometric forward kinematics solved via Levenberg–Marquardt, and task-space control mapped via the Jacobian to joint commands.
  \item an experimental comparative static-stiffness study isolating the effect of universal vs.~spherical base joints, confirming markedly higher torsional stiffness about the vertical axis without degrading in-plane stiffness - critical to suppress twist and vibration under tool loads.
  \item an experimental workspace-scale positioning accuracy study with total-station ground truth, demonstrating millimeter-level closed-loop convergence across heights and spans, and accurate forward kinematics.
  \item an experimental open-loop ceiling drilling study establishing task feasibility by drilling a 15-hole pattern with \SI{4.5}{\milli\meter} maximum relative hole-position error.
\end{itemize}

Tripody deliberately trades workspace volume and full 6-DoF dexterity for lightweight deployment and stable quasi-static operation over a compact footprint. It is best suited to high-reach ceiling tasks that require accurate tool-point positioning, such as drilling, anchoring, screwing and marking, as well as process tasks with modest pose demands such as grinding or sanding. Manipulator-based systems remain preferable when tasks require large lateral coverage, reach-around in clutter, or substantial tool orientation freedom and complex tool trajectories, i.e., full 6-DoF end-effector pose control.

\section{RELATED WORK}\label{sec:related_work}
For interior finishing tasks on construction sites, diverse robotic designs have been explored for operations on ceilings. These include aerial drones, climbing robots, and ground-based manipulators, each trading off reachability, payload capacity, stiffness, and deployment complexity.

Aerial drones offer virtually unrestricted workspace and high agility, and have demonstrated brick-stacking for architectural installations~\cite{augugliaro_flight_2014}, aerial additive manufacturing of cementitious materials~\cite{zhang_aerial_2022}, precise ceiling layout and marking~\cite{lanegger2023chasing}, and wall-perching drilling~\cite{dautzenberg_perching_2023}. Indoor deployment for finishing work is nonetheless impeded by rotor-induced downdrafts that resuspend dust and debris, degrading visibility, posing respiratory risks, and spreading contaminants across the workspace.

Climbing robots maintain close surface contact and generate minimal airflow disturbance, with examples including magnetic-footed legged robots for steel frameworks~\cite{hong_agile_2022}, suction-footed legged robots for smooth walls~\cite{hernando_romerin_2022}, and vacuum-driven crawlers~\cite{hausbots_patent}. In practice, however, ceilings are often cluttered with pre-installed elements (e.g., piping, ducts, fixtures), which obstruct traversal and severely restrict accessible work areas.

Most ground-based systems for interior construction tasks are serial manipulators mounted on wheeled or tracked bases. Such platforms have been used for overhead   drilling~\cite{jaibot_patent,kindle_enhancing_2025,baubot_patent}, drywall finishing~\cite{canvas_app_patent}, spray-painting~\cite{asadi_pictobot_2018}, cooperative timber assembly~\cite{alexi_cooperative_2024}, robotic concrete formwork~\cite{hack_structural_2020}, and modular multi-task operation including wall drilling~\cite{rossini_concert_2026}. Ceiling-reaching serial-chain systems are heavy - often hundreds of kilograms - with payload-to-weight ratios well below \SI{10}{\percent}, which limits on-site deployability and mobility.

Parallel robots enable comparatively lightweight yet stiff structures by distributing loads across multiple limbs~\cite{merlet_parallel_2006}. Representative families include: (i) the Stewart platform (six prismatic legs) providing full 6-DoF positioning, widely used in flight simulation~\cite{pradipta_development_2013} and telescope subsystems~\cite{su_application_2000}; (ii) cable-driven parallel robots that replace compressive links with cables held in positive tension, reducing moving mass and enabling large workspaces - used for stadium cameras~\cite{passarini_dynamic_2019} and proposed for building-envelope installation~\cite{iturralde_cable-driven_2022,Liu2025CDPR}; and (iii) delta robots employing lightweight parallelogram chains that kinematically constrain end-effector orientation, widely adopted for high-speed pick-and-place in assembly and packaging~\cite{ABB_IRB360_FlexPicker_page}.

For lightweight system design, platforms that realize full 3-DoF positioning with three connections between the base and end-effector are particularly attractive. For example, 3-U\underline{P}U architectures are theoretically capable of pure translation but are notoriously sensitive to manufacturing tolerances and alignment, complicating practical realization~\cite{merlet_parallel_2006}. In contrast, \threeRPS{} and \threeSPR{} families are more practical and have seen diverse uses, including wave compensation on ships~\cite{zhan_novel_2020}, self-levelling camera tripods~\cite{Edelkrone_Tripod_X}, and force/torque sensing~\cite{bi_stiffness_2017}. In tall, narrow configurations - i.e., when the base is small relative to the working height - the long legs behave like cantilevers, reducing Cartesian stiffness, with the most pronounced degradation in torsion. This limitation is particularly acute in ceiling operations subject to tool-induced loads - such as overhead drilling and screw-driving - where inadequate torsional stiffness causes twist, chatter, and accuracy degradation. This trend is captured by stiffness/compliance models for lower-mobility parallel robots with flexible limbs~\cite{pashkevich_stiffness_2009}.

Leveraging elasticity to desensitize assemblies to geometric errors - so-called elastic averaging - is a well-established design principle in precision mechanisms~\cite{awtar_elastic2010}. In parallel micro-robots, this is commonly realized by replacing kinematic joints with flexures~\cite{cui_development_2021,gallagher_characterization_2018}, or by relying on structural compliance of the mechanism itself rather than only the joints~\cite{pashkevich_stiffness_2009,ma_evaluation_2024}.

Taken together, this review motivates a lightweight, wheeled, and deployable 3-DoF parallel platform that maintains high torsional stiffness at tall, slender aspect ratios without the mass and deployment overhead of heavy serial manipulators. Accordingly, we replace the spherical base joints of a \threeSPR{} parallel robot with universal joints, intentionally removing one passive rotational DoF per leg; the induced overconstraint is absorbed via elastic averaging. We instantiate this idea in \emph{Tripody}, a \SI{33}{\kilo\gram} \threeSPR{}-like parallel robot with predominantly translational motion and a modular end-effector. In \Cref{sec:system_design}, we detail the system design and control architecture; \Cref{sec:experiments} then quantifies torsional stiffness and positioning accuracy, and validates task-level feasibility through ceiling-drilling experiments.

\section{SYSTEM DESIGN}\label{sec:system_design}
We first give an overview of the system (\Cref{sec:system_overview}), then detail the linear actuator design (\Cref{sec:linear_actuator_design}), followed by the implications of using universal joints (\Cref{sec:compatibility_map}), and finally the control architecture (\Cref{sec:control_architecture}).

\begin{figure}[t]
\includegraphics[width=\linewidth, trim=0.5cm 0cm 0.7cm 0cm, clip]{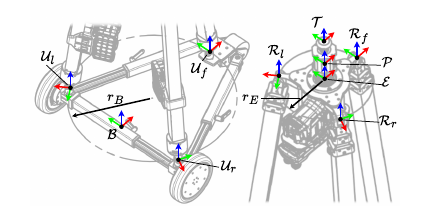}
\caption{Reference frames used in this work.}
\label{fig:frames_overview}
\end{figure}

In the following, scalars are denoted by regular letters (e.g. $a$, $A$), vectors by bold lowercase (e.g. $\mathbf{a}$), and matrices by bold uppercase (e.g. $\mathbf{A}$). We use $\T{A}{B} \in \SEthree$ to denote the rigid transform that maps coordinates from frame $\mathcal{B}$ to frame $\mathcal{A}$, with $\pospart{\T{A}{B}} = \p{A}{B} \in \Rthree$ and $\rotpart{\T{A}{B}} = \R{A}{B} \in \SOthree$ denoting the translation and rotation parts, respectively. We adopt right-multiplicative composition, i.e., $\T{A}{C} = \T{A}{B}\,\T{B}{C}$. With this convention, the action of a rigid transform on a point, denoted by $\odot$, is $\p{A}{C} =\T{A}{B}\!\odot\!\p{B}{C}=\R{A}{B}\,\p{B}{C}+\p{A}{B}$. The reference frames used throughout are shown in \Cref{fig:frames_overview}.
In this work, we use the term \textit{task-space} to refer to the 3-DoF Cartesian position of the end-effector.

\subsection{System Overview}\label{sec:system_overview}
Tripody (\Cref{fig:tripody}) is a wheeled 3-DoF parallel robot whose kinematic chains resemble a classic \threeSPR{} configuration. Each of the three identical legs comprises a prismatic actuator connected to the mobile base by a universal joint and to the end-effector by a revolute joint. By varying the three prismatic extensions, the end-effector motion is predominantly translational in $\mathbb{R}^3$, with small parasitic rotations.

A deliberate departure from a conventional \threeSPR{} architecture is the use of universal rather than spherical joints at the base, removing rotation about each joint’s vertical axis without altering the intended 3-DoF task-space. The substitution introduces a small, workspace-dependent kinematic incompatibility that we deliberately accommodate through distributed structural compliance (elastic averaging). The practical implications of this joint selection are discussed in \Cref{sec:compatibility_map} and empirically characterized in \Cref{sec:static_stiffness_experiments}.

Built primarily from aluminium with a total mass of \SI{33}{\kilo\gram}, Tripody can be carried by two people, making it highly mobile on construction sites. The telescopic base is approximately equilateral and can be adjusted manually from a transport footprint with \SI{57}{\centi\meter} side length to \SI{80}{\centi\meter}, increasing static stability and expanding the usable horizontal workspace at height. The system supports a continuous payload of \SI{32}{\kilo\gram} (\Cref{sec:linear_actuator_design}), yielding a payload-to-weight ratio of nearly $1$, and reaches nominal end-effector speeds of \SI{1.3}{\meter\per\second} horizontally and \SI{0.20}{\meter\per\second} vertically.

The reachable workspace is primarily limited by static stability under quasi-static operation and therefore depends on the mounted end-effector. In the present prototype equipped with the drilling end-effector, we constrain planar motion to an equilateral triangle with side length \SI{50}{\centi\meter}, ensuring the combined center of mass remains within the support polygon. To maintain this stability margin during motion, we cap horizontal accelerations such that inertial effects do not compromise stability. The vertical reach is set by the telescopic linear actuators (\Cref{sec:linear_actuator_design}) and spans from \SI{1.7}{\meter} to \SI{3.4}{\meter}. Within this workspace, deformations remain in the elastic regime (\Cref{sec:compatibility_map}).

These workspace limits assume operation on a firm, approximately planar support surface, as typically expected for indoor ceiling tasks, and quasi-static interaction forces; on uneven or compliant flooring, or near the stability boundary at extended heights, stability margins and positioning accuracy may degrade due to base tilt and lateral tool loads.

The end-effector interface is modular. In \Cref{fig:tripody}, a drilling end-effector with a mass of \SI{4.5}{\kilo\gram} is depicted, allowing ceiling work up to \SI{3.5}{\meter} owing to the tool's standoff.

Because the base is small relative to the working height, small changes in the leg-length map to larger in-plane motion. We quantify this with the planar sensitivity \mbox{$s := \lVert \Delta \mathbf{w}_{xy} \rVert / |\Delta l|$} (with $l$ the actuator length and $\mathbf{w}_{xy}$ the end-effector position in the $xy$-plane), which rises from about $2.8$ at \SI{1.7}{\meter} to about $5.9$ at \SI{3.4}{\meter}. Practically, this implies actuator length must be controlled at least $\sim\!6\times$ finer than the targeted $xy$ positioning accuracy.


\subsection{Linear Actuator Design}\label{sec:linear_actuator_design}
\begin{figure}[t]
\includegraphics[width=\linewidth, trim=0.35cm 0cm 0.3cm -0.05cm, clip]{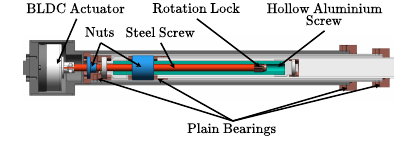}
\caption{Section view of one custom linear actuator, composed of an integrated BLDC motor and a two-stage trapezoidal lead-screw transmission.}
\label{fig:column_sketch}
\end{figure}

\Cref{fig:column_sketch} shows a section view of one leg’s custom prismatic joint. Each leg is driven by a MyActuator \mbox{RMD-L-7015-23T}~\cite{myactuator_l7015} integrated BLDC actuator, directly coupled to a two-stage lead-screw transmission. Both stages use trapezoidal threads; the first screw is steel, the second is a hollow aluminum screw. The corresponding nuts are made of \ac{pom}. A keyed sleeve locks the two coaxial screws in rotation (co-rotate) while allowing axial sliding, enabling relative translation between them. Each stage has a \SI{10}{\milli\meter} lead, resulting in an effective \SI{20}{\milli\meter} linear travel per actuator revolution. Linear guidance relies on \ac{pa12} plain bearings with clearance adjustable via set screws, tuned to minimize backlash. 

Static drive parameters were identified with a two-point vertical load test: two known masses (no load and \SI{5}{\kilo\gram}) were attached at the actuator tip and the steady-state motor torques were recorded. We model the drive force as $F_d(\tau) = \frac{2\pi e}{L}\,\tau$ with transmission efficiency $e$ and lead $L$. With drive velocity $\dot{q}$ and Coulomb threshold $F_c$, the net axial force is
\begin{equation}\label{eq:friction_model}
F(\tau,\dot q)=
\begin{cases}
0 & \dot q = 0 \land|F_d(\tau)| \le F_c \\
F_d(\tau) \!-\! \mathrm{sgn}(\dot q)\,F_c & \text{otherwise}
\end{cases}
\end{equation}
resulting in $e=0.24$ and $F_c=\SI{30}{\newton}$. Let $\tau_n=\SI{1.0}{\newton\meter}$ and $\tau_p=\SI{4.0}{\newton\meter}$ denote the nominal and peak motor torques. Since Coulomb friction opposes lifting, the nominal and peak lifting forces are \mbox{$F_d(\tau_n)-F_c=\SI{44}{\newton}$} and \mbox{$F_d(\tau_p)-F_c=\SI{266}{\newton}$}, respectively. For static payload support, Coulomb friction resists back-driving and thus adds to the holding force; distributing $F_d(\tau_n)+F_c$ across three legs gives a payload of $m_\mathrm{payload}=\frac{3}{g}\bigl(\tfrac{2\pi e}{L}\tau_n+F_c\bigr)=\SI{32}{\kilo\gram}$.

\subsection{Kinematic Compatibility Map}\label{sec:compatibility_map}
\begin{figure}[t]
  \centering
  \includegraphics[width=\linewidth]{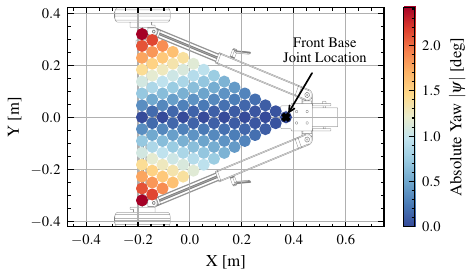}
  \caption{Virtual yaw $|\psi|$ about the front base joint's vertical axis that a virtual spherical joint would require for end-effector positions in the base-frame $xy$-plane at \SI{1.7}{\meter} height, computed via the Tait-Bryan decomposition in \Cref{eq:tait_bryan} to quantify kinematic incompatibility.}
  \label{fig:spherical_yaw_plot_low}
\end{figure}

In this work, we replace the base spherical joints of a conventional 3-SPR architecture with universal joints. This removes yaw about each base joint's vertical axis ($z$) while preserving the intended 3-DoF translational task space. To quantify the resulting kinematic incompatibility, we compute, for each sampled end-effector position, the yaw angle $\psi$ that the base joint would require if modeled as a virtual spherical joint. This yaw captures the rotational mismatch introduced by the universal-joint constraint and is accommodated by elastic deformation of the structure.

To compute $\psi$, we express the virtual joint orientation with a ZYX Tait–Bryan decomposition:
\begin{equation}\label{eq:tait_bryan}
\mathbf{R}(\psi,\theta,\phi)=\mathbf{R}_z(\psi)\,\mathbf{R}_y(\theta)\,\mathbf{R}_x(\phi)
\end{equation}
where $\mathbf{R}_i(x)$ denotes a rotation by angle $x$ about axis $i$, and $\psi,\theta,\phi$ denote yaw, pitch, and roll, respectively.

\Cref{fig:spherical_yaw_plot_low} maps $|\psi|$ for the front base joint at an end-effector height of \SI{1.7}{\meter}. The worst case is \SI{2.4}{\degree} (when the end-effector lies above one of the other base joints). At \SI{3.4}{\meter} height, the worst case decreases to roughly \SI{0.6}{\degree}.

Because yaw at the base is blocked, the kinematically required $\psi$ is realized as elastic twist of the structure; the small values observed ($\leq\SI{2.4}{\degree}$) indicate limited compatibility-induced deformation.

\subsection{Control Architecture}\label{sec:control_architecture}
\begin{figure}[t]
\includegraphics[width=\linewidth, trim=3.0cm 0cm 2.8cm 0cm, clip]{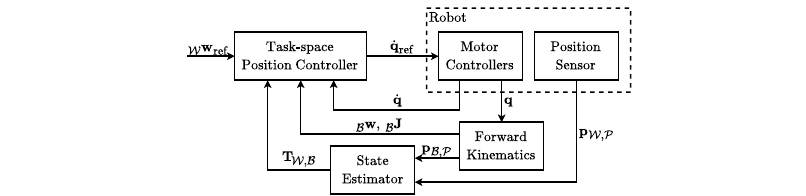}
\caption{Block diagram of the control architecture employed to position the end-effector of Tripody.}
\label{fig:control_architectures}
\end{figure}

The closed-loop end-effector position controller runs at \SI{200}{\hertz} and is illustrated in \Cref{fig:control_architectures}. Our goal is to regulate the tool position ${}_\mathcal{W}\mathbf{w} \!=\! \p{W}{T}$ of a user-defined tool frame $\mathcal{T}$ to a reference ${}_\mathcal{W}\mathbf{w}_\mathrm{ref}$. A high-accuracy position sensor (e.g. a total-station as in~\cite{kindle_enhancing_2025}) provides the position $\p{W}{P}$ of a frame $\mathcal{P}$ colocated near the tool frame $\mathcal{T}$, expressed in the world frame $\mathcal{W}$. We denote joint positions and velocities by $\mathbf{q}$ and $\dot{\mathbf{q}}$; the end-effector position and Jacobian in the base frame $\mathcal{B}$ by ${}_\mathcal{B}\mathbf{w} \!=\! \mathbf{p}_{\mathcal{B},\mathcal{E}}$ and ${}_\mathcal{B}\mathbf{J}$; the sensor position in the base frame by $\p{B}{P}$; the reference joint velocity by $\dot{\mathbf{q}}_\mathrm{ref}$; and the base pose in the world frame by $\T{W}{B}$. All controller gains were empirically tuned on the prototype for fast, stable convergence with minimal overshoot.

\subsubsection{Forward Kinematics}\label{sec:forward_kinematics}
Forward kinematics for parallel robots is generally involved, and only special architectures admit analytic closed-form solutions~\cite{merlet_parallel_2006}. For a general \threeSPR{} or \threeRPS{} manipulator, the forward kinematics reduces to a polynomial system with generic algebraic degree~16~\cite{schadlbauer_3-rps_2014} and is solved numerically in practice~\cite{wang_fast_2018,merlet_parallel_2006}. Although Tripody uses a U\underline{P}R morphology, we employ a classical \threeSPR{} kinematic model that treats the base universal joint as spherical at the kinematic level and absorbs structural compliance into the residuals. We estimate the end-effector pose $\T{B}{E}\in \SEthree$ via damped least squares (Levenberg–Marquardt) by minimizing
\begin{equation}
\begin{aligned}
\hat{\mathbf{T}}\!_{\mathcal{B},\mathcal{E}} \!=\! \argmin_{\T{B}{E}}
\left\lVert 
    \!\left[\mathbf{c}^{(f)}(\T{B}{E}),
    \mathbf{c}^{(l)}(\T{B}{E}),
    \mathbf{c}^{(r)}(\T{B}{E})\right]\!
\right\rVert_2^2
\end{aligned}
\end{equation}
where the per-leg residuals $\mathbf{c}^{(i)}$ with $i\in\{f,l,r\}$ indexing the front, left, and right legs, enforce that the universal-joint location, expressed in the $i$-th revolute-joint frame $\mathcal{R}_i$, lies in the joint’s rotation plane at a radial distance equal to the measured prismatic extension $l_i$
\begin{equation}
\begin{aligned}
\mathbf{c}^{(i)}(\T{B}{E}) \!=\!
\left[
  l_i - \left\lVert \mathbf{p}^{(x,z)}_{\mathcal{R}_i,\mathcal{U}_i}\!\left(\T{B}{E}\right) \right\rVert_2,
  \;\mathrm{p}^{(y)}_{\mathcal{R}_i,\mathcal{U}_i}\!\left(\T{B}{E}\right)
\right]\\
\mathbf{p}_{\mathcal{R}_i,\mathcal{U}_i}\!\left(\T{B}{E}\right)
= \left(\T{B}{E}\,\Traw{\mathcal{E}}{\mathcal{R}_i}\right)^{-1} \!\odot\, \praw{\mathcal{B}}{\mathcal{U}_i}
\end{aligned}
\end{equation}
with $\mathbf{T}_{\mathcal{E},\mathcal{R}_i}$ the pose of the $i$-th revolute joint in the end-effector frame and $\praw{\mathcal{B}}{\mathcal{U}_i}$ the position of the $i$-th universal joint in the base frame. Here, $\praw{\mathcal{R}_i}{\mathcal{U}_i}=[p_x,\,p_y,\,p_z]^\top$, $\mathbf{p}^{(x,z)}_{\mathcal{R}_i,\mathcal{U}_i}=[p_x,\,p_z]^\top$, and $\mathrm{p}^{(y)}_{\mathcal{R}_i,\mathcal{U}_i}=p_y$.

\subsubsection{State Estimator}\label{sec:state_estimator}
To transform the setpoint from world to base frame, we estimate the base pose at time $k$, \mbox{$\boldsymbol{X}_k=\T{W}{B}^{(k)}\in\SEthree$}, using an \ac{iekf} on $\SEthree$~\cite{gtsam}. Assuming instantaneous planar motion of the mobile base, the body-frame twist input (ordered as $[\omega_x,\omega_y,\omega_z,\,v_x,v_y,v_z]$) is $\boldsymbol{u}_{k-1}=[0,\,0,\,\omega^{(z)}_{k-1},\,v^{(x)}_{k-1},\,0,\,0]^\top$. The discrete-time left-invariant propagation and the total-station position update are modeled as
\begin{equation}
\begin{aligned}
    \boldsymbol{X}_{k+1} &= \boldsymbol{X}_{k}\,
      \mathrm{Exp}\!\big(\boldsymbol{u}_{k}\,\Delta t\big)\,
      \mathrm{Exp}\!\big(\boldsymbol{w}_{k}\big), 
    & \boldsymbol{w}_{k}&\sim\mathcal{N}(\boldsymbol{0},\mathbf{Q})\\
    \boldsymbol{z}_k &= \boldsymbol{X}_k \odot \p{B}{P}^{(k)} + \boldsymbol{v}_k,
    & \boldsymbol{v}_k&\sim\mathcal{N}(\boldsymbol{0},\mathbf{R})
\end{aligned}
\end{equation}
where $\mathrm{Exp}:\mathfrak{se}(3)\!\to\!\SEthree$ is the group exponential, $\Delta t$ is the delta time from the last state, $\p{B}{P}^{(k)}$ is the forward-kinematics position of the sensor frame $\mathcal{P}$ expressed in $\mathcal{B}$ at time $k$, $\mathbf{Q}$ is the covariance of body-frame twist noise in $\mathfrak{se}(3)$ units, and $\mathbf{R}$ is the covariance of the position measurement $\boldsymbol{z}_k \in \mathbb{R}^3$ in world coordinates.

\subsubsection{Task-space Position Controller}\label{sec:position_controller}
The task-space position controller implements a PD feedback law of the form
\begin{equation}\label{eq:task_space_controller}
\begin{aligned}
{}_\mathcal{B}\dot{\mathbf{w}}_{\mathrm{ref}}
&= \mathbf{K}_P^{(\mathbf{w})}\bigl({}_\mathcal{B}\mathbf{w}_{\mathrm{ref}} - {}_\mathcal{B}\mathbf{w}\bigr)
- \mathbf{K}_D^{(\mathbf{w})}\,{}_\mathcal{B}\dot{\mathbf{w}}\\
&{}_\mathcal{B} \mathbf{w}_\mathrm{ref} = \T{W}{B}^{-1} \odot {}_\mathcal{W}\mathbf{w}_\mathrm{ref}\\
\dot{\mathbf{q}}_\mathrm{ref} &= {}_\mathcal{B}\mathbf{J}^{-1}\,{}_\mathcal{B}\dot{\mathbf{w}}_\mathrm{ref}
\end{aligned}
\end{equation}
where $\mathbf{K}_P^{(\mathbf{w})},\mathbf{K}_D^{(\mathbf{w})}\in\mathbb{R}^{3\times3}$ are the Cartesian proportional and derivative gain matrices, and ${}_\mathcal{B}\mathbf{J}$ denotes the end-effector Jacobian. Using diagonal gains with identical entries and the differential kinematics ${}_\mathcal{B}\dot{\mathbf{w}} = {}_\mathcal{B}\mathbf{J}\,\dot{\mathbf{q}}$, \Cref{eq:task_space_controller} simplifies to
\begin{equation}
\dot{\mathbf{q}}_\mathrm{ref}
= K_P^{(\mathbf{w})}\,{}_\mathcal{B}\mathbf{J}^{-1}\bigl({}_\mathcal{B}\mathbf{w}_{\mathrm{ref}} - {}_\mathcal{B}\mathbf{w}\bigr)
- K_D^{(\mathbf{w})}\,\dot{\mathbf{q}}
\end{equation}
with $K_P^{(\mathbf{w})}, K_D^{(\mathbf{w})} \in \mathbb{R}$ denoting the scalar proportional and derivative gains.

\subsubsection{Motor Controllers}\label{sec:motor_controllers}
Velocity and current regulation on each actuator are implemented via a cascade of two PI loops embedded in the motor-drive electronics, running at \SI{5}{\kilo\hertz} (velocity loop) and \SI{15}{\kilo\hertz} (current loop). Denoting the measured joint/motor speed by $\dot{q}$ and the phase current by $I$, the outer velocity loop computes the current reference $I_{\mathrm{ref}}$, and the inner current loop the voltage command $U_{\mathrm{ref}}$:
\begin{equation}
\begin{aligned}
I_{\mathrm{ref}} &=
K_{P}^{(\dot{q})}\bigl(\dot{q}_{\mathrm{ref}} - \dot{q}\bigr)
+ K_{I}^{(\dot{q})}\!\int\!\bigl(\dot{q}_{\mathrm{ref}} - \dot{q}\bigr)\,\mathrm{d}t\\
U_{\mathrm{ref}} &=
K_{P}^{(I)}\bigl(I_{\mathrm{ref}} - I\bigr)
+ K_{I}^{(I)}\!\int\!\bigl(I_{\mathrm{ref}} - I\bigr)\,\mathrm{d}t
\end{aligned}
\end{equation}
where $K_{P}^{(\dot{q})}$, $K_{I}^{(\dot{q})}$, $K_{P}^{(I)}$, and $K_{I}^{(I)}$ are the proportional and integral gains for the velocity and current loops, respectively.

\section{EXPERIMENTAL VERIFICATION}\label{sec:experiments}
This section validates the proposed mechanical design and control architecture through two task-agnostic characterization experiments and a representative system-level demonstration. First, we quantify static stiffness and show the torsional-stiffness benefit of replacing spherical with universal base joints (\Cref{sec:static_stiffness_experiments}). Second, we evaluate closed-loop tool-positioning accuracy using total-station feedback and compare forward-kinematics residuals against a serial construction robot baseline across representative heights and spans (\Cref{sec:accuracy_experiments}). Finally, we validate task-level ceiling-drilling feasibility by evaluating relative hole-pattern accuracy under realistic tool-material interaction, including open-loop positioning, compliance, and drilling-induced deviations (\Cref{sec:drilling_experiments}). In all experiments, the two actuated wheels are commanded to fixed angles and held with high stiffness via position control, so the base does not roll during measurement.

\subsection{Static Stiffness Experiments}\label{sec:static_stiffness_experiments}
\begin{figure}[t]
  \centering
  \vspace{2mm}
  \includegraphics[width=\linewidth, trim=0cm 0cm 0cm 0cm, clip]{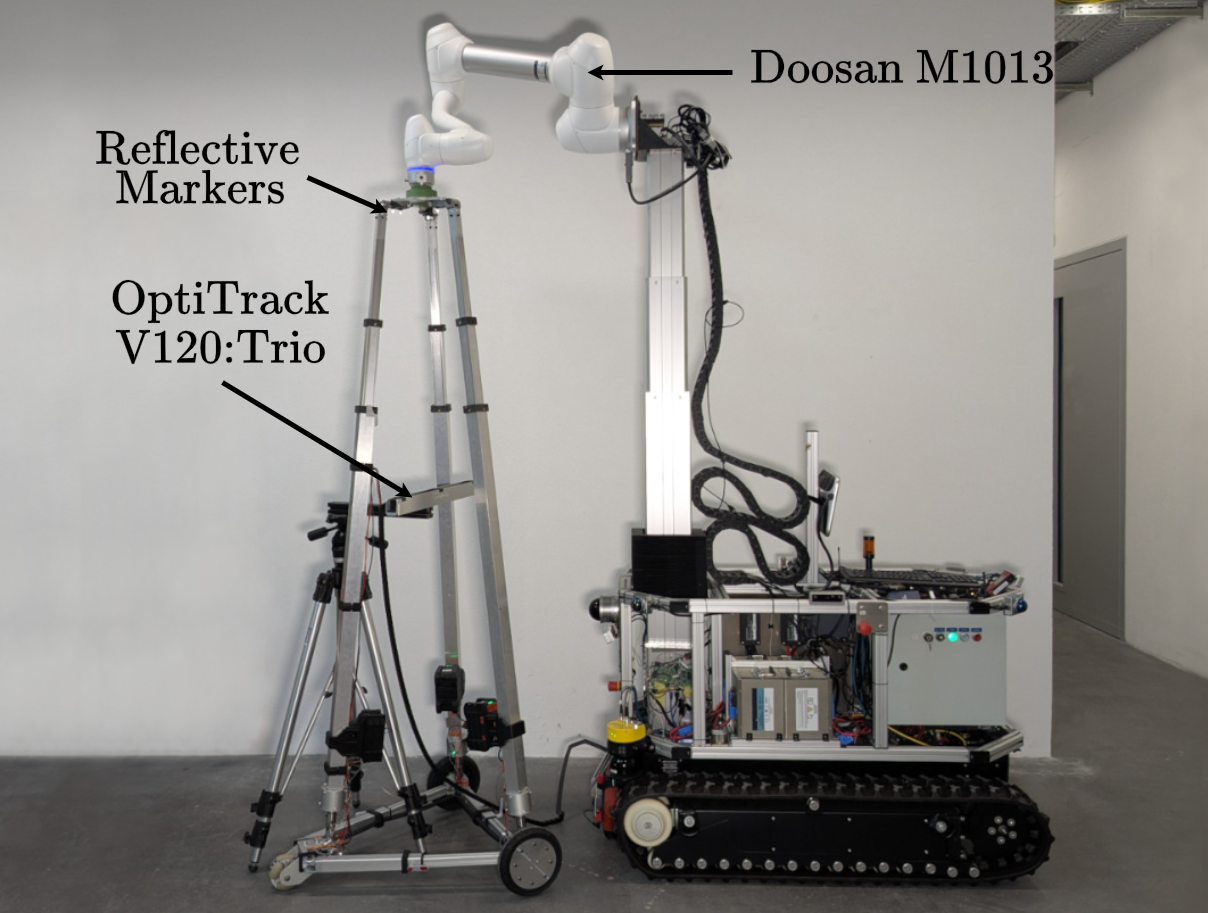}
  \caption{Stiffness measurement setup. An external wrench is applied to Tripody’s end-effector with a \textit{Doosan M1013}; the resulting displacement is measured with an \textit{OptiTrack V120:Trio}.}
  \label{fig:stiffness_experiment}
\end{figure}

\begin{figure*}[t!] \centering \includegraphics[width=\textwidth]{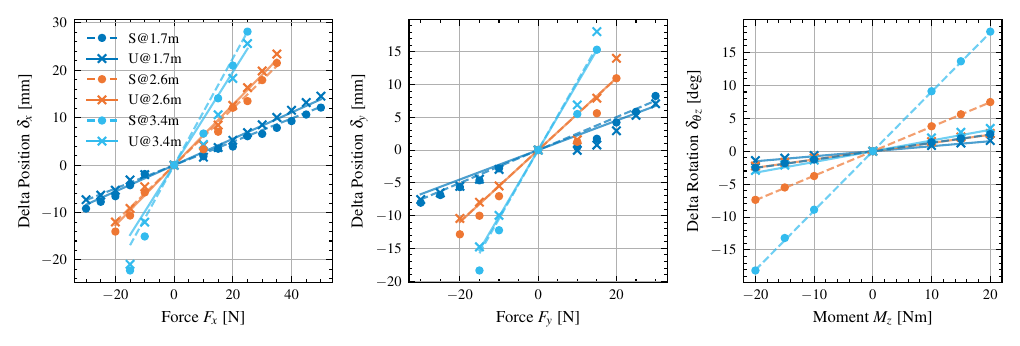} \caption{In-axis Cartesian stiffness at the workspace center. Force–displacement along $x$ and $y$ and moment–rotation about $z$ are shown for three tool heights and both the Universal (U) and Spherical (S) joint configurations. Lines depict least-squares fits to the measurements.} \label{fig:stiffness_plots} \end{figure*}

For tasks such as overhead drilling and other on-tool operations, end-effector stiffness - especially in torsion - governs how much the structure twists or deflects under external tool loads; higher stiffness suppresses oscillations and supports better accuracy. 
Representative commercial tool specifications suggest that quasi-static tool reaction torques can range from a few to several tens of \si{\newton\meter}, e.g. a few \si{\newton\meter} for concrete grinding/polishing tools, around \SIrange{10}{15}{\newton\meter} for compact drills, around \SIrange{20}{25}{\newton\meter} for hand-held core drills, and around \SI{40}{\newton\meter} for compact screwdriving/driver-drill tools.\footnote{Order-of-magnitude estimates from representative tool specifications, using $\tau \approx P/\omega$, where $\tau$ is torque, $P$ is power, and $\omega$ is spindle speed.} This makes torsion about the vertical axis a limiting case.
Motivated by this, we compare our universal-joint base design against a spherical-joint variant via quasi-static, in-axis stiffness identification at three end-effector heights.

As depicted in \Cref{fig:stiffness_experiment}, for both joint configurations we apply external wrenches at the end-effector at heights of \SI{1.7}{\meter}, \SI{2.6}{\meter}, and \SI{3.4}{\meter} using a \textit{Doosan M1013} with joint torque sensors and measure the resulting quasi-static pose change with an \textit{OptiTrack V120:Trio} motion-capture system.

Stiffness is identified by fitting a straight line to the measurements and taking the inverse slope. Concretely,
\begin{equation}
\begin{aligned}
    \delta_i \approx m_i F_i \rightarrow k_{i} = \frac{1}{m_i} 
\end{aligned}
\end{equation}
where $\delta_i$ denotes the translational displacement ($x$ or $y$) or the rotational displacement ($\theta_z$), $F_i$ the applied force ($F_x$ or $F_y$) or moment ($M_z$), $m_i$ the fitted slope (compliance), and $k_i$ the resulting in-axis stiffness (least-squares fit with zero intercept).

\Cref{fig:stiffness_plots} depicts the in-axis Cartesian stiffness analysis for the linear $x$ and $y$ axes as well as rotational $z$ axis. The universal-joint configuration yields similar stiffness in $x$ and $y$ and a markedly higher torsional stiffness about $z$: $k_{\theta_z}$ increases by \SI{67}{\percent} at \SI{1.7}{\meter} (from \SI{7.9}{\newton\meter\per\deg} to \SI{13.1}{\newton\meter\per\deg}), by \SI{196}{\percent} at \SI{2.6}{\meter} (from \SI{2.7}{\newton\meter\per\deg} to \SI{8.0}{\newton\meter\per\deg}), and by \SI{454}{\percent} at \SI{3.4}{\meter} (from \SI{1.1}{\newton\meter\per\deg} to \SI{6.1}{\newton\meter\per\deg}). Cross-axis effects were negligible over the tested range.

At maximum height, tool reaction torques of \SI{10}{\newton\meter} to \SI{30}{\newton\meter} correspond to quasi-static yaw deflections of \SI{1.6}{\degree} to \SI{4.9}{\degree}, indicating feasibility for moderate-torque ceiling operations in the range of compact drilling and screwdriving.

\subsection{Positioning Accuracy Experiments}\label{sec:accuracy_experiments}
\begin{figure}[t!]
\includegraphics[width=\linewidth, trim=0cm 0.55cm 0cm 0.6cm, clip]{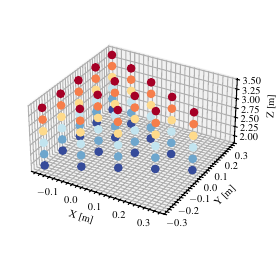}
\caption{A total of 90 setpoints arranged on 2D uniform triangular grids within the workspace at six heights between \SI{1.9}{\meter} and \SI{3.4}{\meter}.}
\label{fig:setpoint_grid}
\end{figure}

\begin{figure}[t!]
\includegraphics[width=\linewidth]{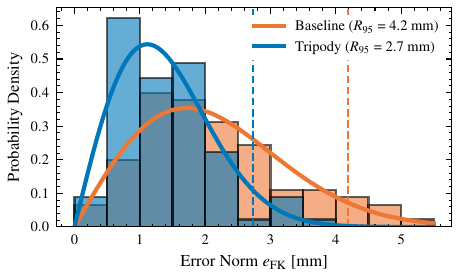}
\caption{Density histograms of the forward-kinematics prism-position residual magnitude $e_{\mathrm{FK}}$~(\Cref{eq:fk_error}) over 90 workspace setpoints. Solid line: Rayleigh fit; dashed line: fitted 95th-percentile threshold.}
\label{fig:fk_fit_histogram}
\end{figure}

We evaluate Tripody’s closed-loop tool-positioning accuracy using a total station as the position sensor, consistent with mm-level site instrumentation in construction~\cite{jaibot_patent,lanegger2023chasing,alexi_cooperative_2024,kindle_enhancing_2025,Liu2025CDPR,ercan_automated_2019}. Specifically, we use a \textit{Hilti PLT 400-2}~\cite{Hilti_PLT_400_2} with a static position RMS of \SI{0.3}{\milli\meter} and a 95th-percentile error of \SI{0.5}{\milli\meter}.

For a total of 90 reference positions distributed across the workspace (\Cref{fig:setpoint_grid}), Tripody converges to each setpoint under closed-loop control with total-station feedback. Positioning error is computed as the Euclidean norm between the measured prism position and its reference,
\begin{equation}\label{eq:positioning_error}
e_p \;=\; \left\lVert {}_\mathcal{W}\mathbf{w}_\textrm{ref} - \p{W}{P}\right\rVert_2
\end{equation}
where we identify the tool frame $\mathcal{T}$ with the prism frame (\emph{i.e.}, $\mathcal{P}\equiv\mathcal{T}$) since we command the prism position directly. Across all setpoints, the closed-loop controller converges with an RMS error of \SI{0.4}{\milli\meter} and a maximum observed error below \SI{0.6}{\milli\meter}. Given the total station noise level, these results suggest that the measured residuals are largely sensor-noise-limited.

We additionally assess the forward-kinematics  model using these data. We fit the base radius $r_B$, end-effector radius $r_E$, prism $z$-offset ${t}^{(z)}_{\mathcal{E},\mathcal{P}}$, and the base pose $\mathbf{T}_{\mathcal{W},\mathcal{B}}$ by least squares, and then evaluate the prism-position residuals
\begin{equation}\label{eq:fk_error}
e_{\mathrm{FK}} \;=\; \left\lVert \mathbf{p}_{\mathcal{W},\mathcal{P}} - \hat{\mathbf{p}}_{\mathcal{W},\mathcal{P}}(\mathbf{q}) \right\rVert_2
\end{equation}
where $\hat{\mathbf{p}}_{\mathcal{W},\mathcal{P}}(\mathbf{q})=\T{W}{B}\odot\p{B}{P}(\mathbf{q})$ is the forward-kinematics-predicted prism position in $\mathcal{W}$, and $\p{B}{P}(\mathbf{q})$ is the corresponding base-frame prediction from the \threeSPR{} model (\Cref{sec:forward_kinematics}).

As a baseline, we repeated the same 90-setpoint pure forward-kinematics evaluation with a robot used in prior work on ceiling-task accuracy~\cite{kindle_enhancing_2025}. The histogram of 3D prediction errors is shown in \Cref{fig:fk_fit_histogram}. Over all points, Tripody's error was at most \SI{3.6}{\milli\meter} with no notable height dependence. A Rayleigh fit to the error norms yields a 95th-percentile of \SI{2.7}{\milli\meter}, compared with \SI{4.2}{\milli\meter} for the serial robot under the same evaluation protocol. This indicates that Tripody remains accurate for quasi-static tasks even when extrapolating from a known end-effector pose via forward kinematics.

\subsection{Drilling Experiments}\label{sec:drilling_experiments}

\begin{figure}[t!]
\centering
\vspace{2mm}
\includegraphics[width=\linewidth, trim=0.0cm 0.5cm 0.0cm 0.5cm, clip]{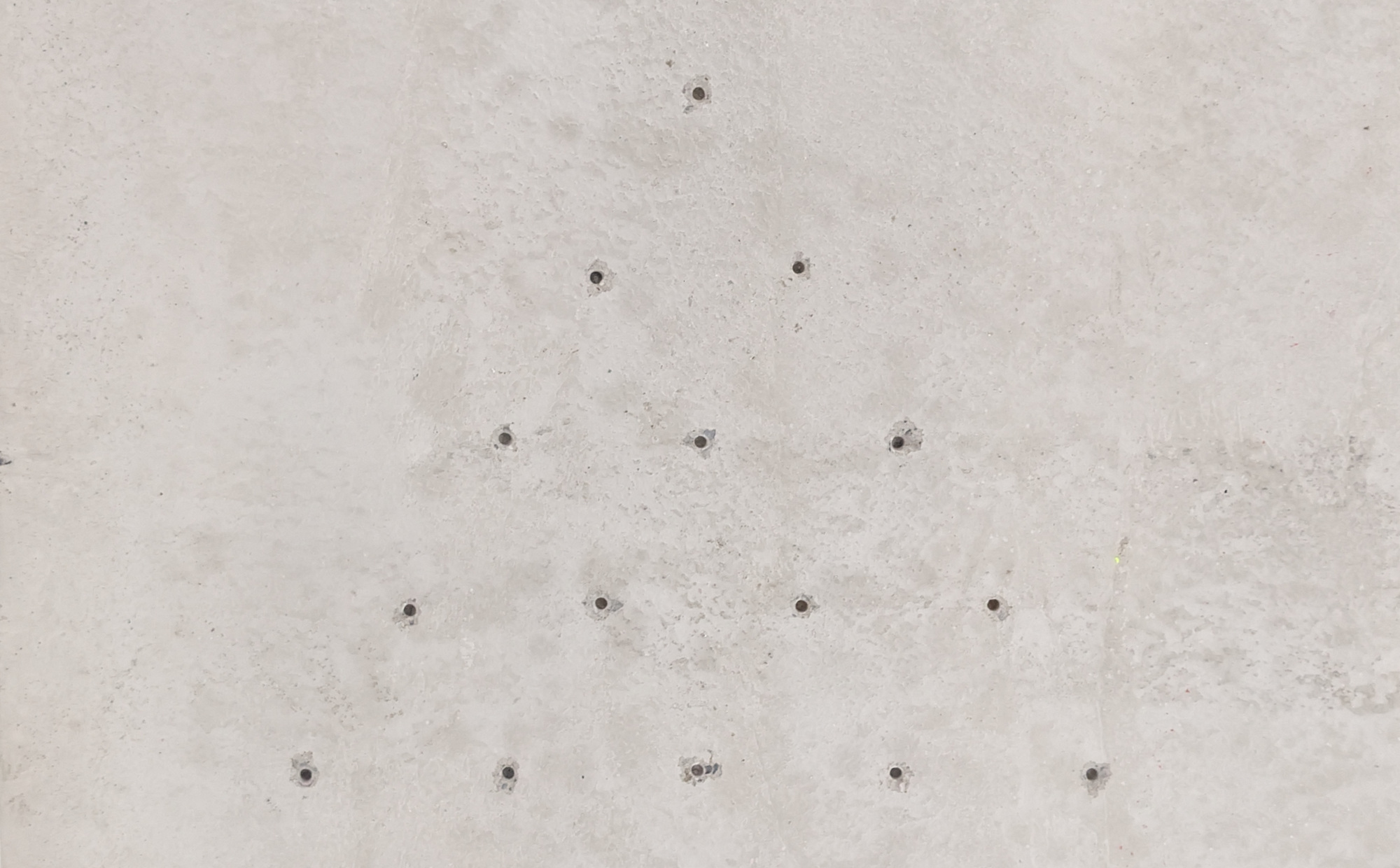}
\caption{Drilled hole pattern produced in the open-loop drilling experiment. Tripody drilled 15 holes (\ensuremath{\varnothing}\SI{8}{\milli\meter}) on a triangular grid with \SI{50}{\centi\meter} side length.}
\label{fig:drilled_holes}
\end{figure}

To evaluate task-level performance under realistic tool–material interaction, we conducted drilling experiments with Tripody equipped with the end-effector drill~(\Cref{fig:tripody}). The employed drill requires the structure to withstand up to \SI{10}{\newton\meter} of applied reaction torque during operation, corresponding to a quasi-static yaw deflection of approximately \SI{1.6}{\degree} at maximum height (\Cref{sec:static_stiffness_experiments}). Throughout the experiment, the system remained stable while drilling under this loading regime. Unlike the accuracy experiments in \Cref{sec:accuracy_experiments}, drilling introduces additional process-dependent effects that directly affect the realized hole location, including play of the drill bit in the chuck and transient "jumping" or walking of the drill at the onset of contact. This experiment therefore evaluates relative hole-pattern accuracy under realistic tool-material interaction, complementing the absolute tool-positioning and forward-kinematics evaluations in \Cref{sec:accuracy_experiments}.

In this experiment, Tripody drilled $15$ holes of diameter \SI{8}{\milli\meter} in open loop at a height of \SI{3.2}{\meter}~(\Cref{fig:drilled_holes}). Target locations were arranged on a triangular grid with side length \SI{50}{\centi\meter}; the robot visited each target sequentially and executed an identical drilling cycle at each site, with an average cycle time of \SI{13}{\second} per hole over the full pattern. To evaluate the local pattern fidelity targeted by this experiment, we report relative hole-center errors after factoring out a global rigid alignment in 2D, computed via the Kabsch algorithm~\cite{Kabsch1976}. This removes the common translation and rotation of the drilled pattern while retaining hole-to-hole errors from open-loop positioning and drilling-induced deviations, thereby quantifying local pattern consistency. Consequently, this experiment does not directly evaluate absolute drilled-hole accuracy in the world or building frame, which remains a limitation of the present evaluation.

\begin{figure}[t!]
\includegraphics[width=\linewidth]{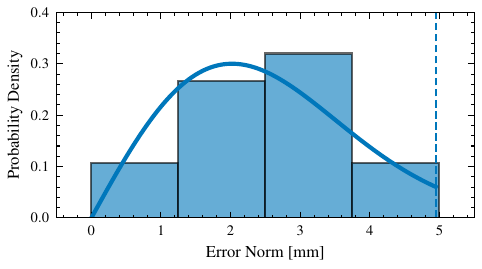}
\caption{Density histogram of the 2D relative hole-position error over $15$ drilled holes. Solid line: Rayleigh fit; dashed line: fitted 95th-percentile threshold.}
\label{fig:drilling_histogram}
\end{figure}

The histogram of 2D relative hole-position errors is shown in \Cref{fig:drilling_histogram}. Across all $15$ holes, the maximum hole-center error after rigid alignment was \SI{4.5}{\milli\meter}. Fitting a Rayleigh distribution to the error magnitudes yields a 95th-percentile error of \SI{4.9}{\milli\meter}, indicating that the drilled pattern remains tightly consistent even in the presence of drilling-process disturbances (e.g., initial bit walking and chuck/bit play).

\section{CONCLUSION}\label{sec:conclusion}
We presented Tripody, a lightweight, wheeled, and field-deployable 3-DoF parallel robot whose U\underline{P}R legs each intentionally remove one passive rotational DoF of a classical \threeSPR{} morphology. The resulting overconstraint is absorbed by small elastic deflections, preserving predominantly translational motion while substantially increasing torsional stiffness - critical for many high-reach ceiling tasks.

A comparative stiffness study between the proposed universal-base design and a classical \threeSPR{} variant shows large torsional gains about $z$ without degrading in-plane stiffness: \SI{67}{\percent} at \SI{1.7}{\meter}, \SI{196}{\percent} at \SI{2.6}{\meter}, and \SI{454}{\percent} at \SI{3.4}{\meter}, with negligible cross-axis coupling.

We proposed a closed-loop estimation-and-control framework tailored to high-accuracy positioning - combining state estimation, forward kinematics, and task-space control - that achieves a maximum observed closed-loop positioning error below \SI{0.6}{\milli\meter}. Pure model extrapolation attains a 95th-percentile position error of \SI{2.7}{\milli\meter} (max \SI{3.6}{\milli\meter}) across the tested workspace.

Finally, we validated task-level feasibility in ceiling drilling with Tripody by executing a 15 hole pattern. After rigid alignment, the relative hole-position errors yield a maximum error of \SI{4.5}{\milli\meter}, indicating consistent pattern execution despite drilling disturbances such as initial bit walking and chuck/bit play. These findings support overconstrained large-scale \threeSPR{} designs as practical and precise for high-reach construction tasks.

\bibliographystyle{IEEEtran}
\bibliography{references} 

@IEEEtranBSTCTL{BSTcontrol,
  CTLuse_forced_etal        = "yes",
  CTLmax_names_forced_etal  = "6",
  CTLnames_show_etal        = "1"
}

@article{rossini_concert_2026,
	title = {{CONCERT}: {A} {Modular} {Reconfigurable} {Robot} for {Construction}},
	volume = {43},
	issn = {1556-4967},
	shorttitle = {{CONCERT}},
	doi = {10.1002/rob.70092},
	language = {en},
	number = {3},
	urldate = {2026-05-06},
	journal = {J. Field Robotics},
	author = {Rossini, Luca and Romiti, Edoardo and Laurenzi, Arturo and Ruscelli, Francesco and Ruzzon, Marco and Covizzi, Luca and Baccelliere, Lorenzo and Carrozzo, Stefano and Terzer, Michael and Magri, Marco and Morganti, Carlo and Lei, Maolin and Bertoni, Liana and Vedelago, Diego and Burchielli, Corrado and Cordasco, Stefano and Muratore, Luca and Giusti, Andrea and Tsagarakis, Nikos},
	year = {2026},
	pages = {1332--1362},
}

@inproceedings{gallagher_characterization_2018,
  author    = {Gallagher, Benjamin B. and Knight, J. Scott and Acton, D. Scott and Smith, Koby Z. and Wolf, Erin and Coppock, Eric and Tersigni, James and Comeau, Thomas and Chonis, Taylor S.},
  title     = {Characterization and calibration of the {James} {Webb} space telescope mirror actuators fine stage motion},
  booktitle = {Proc. SPIE, Space Telescopes and Instrum.},
  volume    = {10698},
  pages     = {106983S},
  year      = {2018},
  doi       = {10.1117/12.2311815},
}

@article{cui_development_2021,
	title = {Development of a 3-{DOF} {Flexible} {Micro}-{Motion} {Platform} {Based} on a {New} {Compound} {Lever} {Amplification} {Mechanism}},
	volume = {12},
	copyright = {http://creativecommons.org/licenses/by/3.0/},
	issn = {2072-666X},
	doi = {10.3390/mi12060686},
	language = {en},
	number = {6},
	urldate = {2025-09-09},
	journal = {Micromachines},
	author = {Cui, Fangni and Li, Yangmin and Qian, Junnan},
	month = jun,
	year = {2021},
	pages = {686},
}

@article{awtar_elastic2010,
    author = {Awtar, Shorya and Shimotsu, Kevin and Sen, Shiladitya},
    title = {Elastic Averaging in
Flexure Mechanisms: A Three-Beam Parallelogram Flexure Case
Study},
    journal = {J. Mech. Robot.},
    volume = {2},
    number = {4},
    pages = {041006},
    year = {2010},
    month = {09},
    issn = {1942-4302},
    doi = {10.1115/1.4002204},
    eprint = {https://asmedigitalcollection.asme.org/mechanismsrobotics/article-pdf/2/4/041006/6623173/041006_1.pdf},
}

@article{pashkevich_stiffness_2009,
	title = {Stiffness {Analysis} of {Overconstrained} {Parallel} {Manipulators}},
	volume = {44},
	issn = {0094114X},
	doi = {10.1016/j.mechmachtheory.2008.05.017},
	number = {5},
	urldate = {2025-09-09},
	journal = {Mech. Mach. Theory},
	author = {Pashkevich, Anatoly and Chablat, Damien and Wenger, Philippe},
	month = may,
	year = {2009},
	pages = {966--982},
}

@article{ma_evaluation_2024,
	title = {Evaluation on {Configuration} {Stiffness} of {Overconstrained} {2R1T} {Parallel} {Mechanisms}},
	volume = {37},
	issn = {2192-8258},
	doi = {10.1186/s10033-024-01045-1},
	number = {1},
	urldate = {2025-09-11},
	journal = {Chin. J. Mech. Eng.},
	author = {Ma, Xuejian and Xu, Zhenghe and Xu, Yundou and Wang, Yu and Yao, Jiantao and Zhao, Yongsheng},
	month = jul,
	year = {2024},
	pages = {62},
}

@inproceedings{Liu2025CDPR,
  author    = {Y. Liu and N. W. Hayes and B. Selvakumar and D. Hun and B. P. Maldonado},
  title     = {Development of a lab-scale prototype of a cable-driven parallel robot for automated installation of prefabricated building envelopes},
  booktitle = {Proc. 4th Workshop Future Construction, IEEE ICRA Workshops},
  year      = {2025},
  note      = {Extended abstract},
  url       = {https://construction-robots.github.io/papers/83.pdf}
}

@misc{ABB_IRB360_FlexPicker_page,
  author       = {{ABB Robotics}},
  title        = {{IRB 360 FlexPicker}},
  note          = {{Product Page}, \url{https://one.robotics.abb.com/en/robots/p/IRB-360} {Accessed}: 2026-05-21}
}

@misc{myactuator_l7015,
  author       = {{MyActuator}},
  title        = {{RMD-L-7015-23T}},
note          = {{Product Page}, \url{https://www.myactuator.com/l-7015-details} {Accessed}: 2026-05-21}
}

@misc{Edelkrone_Tripod_X,
  author       = {{Edelkrone}},
  title        = {{Tripod X}},
  note          = {{Product Page}, \url{https://edelkrone.com/products/tripod-x} {Accessed}: 2026-05-21}
}

@misc{Hilti_PLT_400_2,
  author       = {{Hilti}},
  title        = {{PLT 400-2}},
  note          = {{Product Page}, \url{https://www.hilti.com/r16318483} {Accessed}: 2026-05-21}
}

@book{merlet_parallel_2006,
	address = {Dordrecht},
	edition = {2nd ed},
	series = {Solid mechanics and its applications},
	title = {Parallel robots},
	isbn = {978-1-4020-4132-7},
	language = {en},
	number = {volume 128},
	publisher = {Springer},
	author = {Merlet, Jean-Pierre},
	year = {2006},
}

@article{wang_fast_2018,
	title = {Fast forward kinematics algorithm for real-time and high-precision control of the 3-{RPS} parallel mechanism},
	volume = {13},
	issn = {2095-0241},
	doi = {10.1007/s11465-018-0519-5},
	language = {en},
	number = {3},
	urldate = {2025-08-27},
    journal = {Front. Mech. Eng.},
    author = {Wang, Yue and Yu, Jingjun and Pei, Xu},
	month = sep,
	year = {2018},
	pages = {368--375},
}

@article{schadlbauer_3-rps_2014,
	title = {The 3-{RPS} parallel manipulator from an algebraic viewpoint},
	volume = {75},
	issn = {0094114X},
	doi = {10.1016/j.mechmachtheory.2013.12.007},
	language = {en},
	urldate = {2025-08-27},
	journal = {Mech. Mach. Theory},
	author = {Schadlbauer, J. and Walter, D.R. and Husty, M.L.},
	month = may,
	year = {2014},
	pages = {161--176},
}

@article{zhan_novel_2020,
	title = {A {Novel} {Three}-{SPR} {Parallel} {Platform} for {Vessel} {Wave} {Compensation}},
	volume = {8},
	issn = {2077-1312},
	doi = {10.3390/jmse8121013},
	language = {en},
	number = {12},
	urldate = {2025-09-09},
	journal = {J. Mar. Sci. Eng.},
	author = {Zhan, Yong and Tian, Huichun and Xu, Jianan and Wu, Shaofei and Fu, Junsheng},
	month = dec,
	year = {2020},
	pages = {1013},
}

@inproceedings{pradipta_development_2013,
	title = {Development of a pneumatically driven flight simulator {Stewart} platform using motion and force control},
	doi = {10.1109/AIM.2013.6584085},
	urldate = {2025-09-09},
    booktitle = {Proc. IEEE/ASME Int. Conf. Adv. Intell. Mechatronics (AIM)},
    author = {Pradipta, Justin and Klünder, Mario and Weickgenannt, Martin and Sawodny, Oliver},
	month = jul,
	year = {2013},
	pages = {158--163},
}

@article{su_application_2000,
	title = {The application of the {Stewart} platform in large spherical radio telescopes},
	volume = {17},
	issn = {1097-4563},
	doi = {10.1002/1097-4563(200007)17:7<375::AID-ROB3>3.0.CO;2-7},
	language = {en},
	number = {7},
	urldate = {2025-09-09},
	journal = {J. Robot. Syst.},
	author = {Su, Y. X. and Duan, B. Y.},
	year = {2000},
	pages = {375--383},
}

@article{iturralde_cable-driven_2022,
	title = {Cable-driven parallel robot for curtain wall module installation},
	volume = {138},
	issn = {0926-5805},
	doi = {10.1016/j.autcon.2022.104235},
	urldate = {2025-09-09},
	journal = {Autom. Constr.},
	author = {Iturralde, K. and Feucht, M. and Illner, D. and Hu, R. and Pan, W. and Linner, T. and Bock, T. and Eskudero, I. and Rodriguez, M. and Gorrotxategi, J. and Izard, J. B. and Astudillo, J. and Cavalcanti Santos, J. and Gouttefarde, M. and Fabritius, M. and Martin, C. and Henninge, T. and Nornes, S. M. and Jacobsen, Y. and Pracucci, A. and Cañada, J. and Jimenez-Vicaria, J. D. and Alonso, R. and Elia, L.},
	month = jun,
	year = {2022},
	pages = {104235},
}

@article{passarini_dynamic_2019,
	title = {Dynamic {Trajectory} {Planning} for {Failure} {Recovery} in {Cable}-{Suspended} {Camera} {Systems}},
	volume = {11},
	issn = {1942-4302, 1942-4310},
	doi = {10.1115/1.4041942},
	language = {en},
	number = {2},
	urldate = {2025-09-09},
	journal = {J. Mech. Robot.},
	author = {Passarini, Chiara and Zanotto, Damiano and Boschetti, Giovanni},
	month = apr,
	year = {2019},
	pages = {021001},
}

@inproceedings{bi_stiffness_2017,
	title = {Stiffness analysis and verification of one 3 - {RPS} parallel sensor},
	doi = {10.1109/ICMA.2017.8016031},
	urldate = {2025-09-09},
    booktitle = {Proc. IEEE Int. Conf. Mechatronics Automat. (ICMA)},
    author = {Bi, Liangyu and Jia, Wenchuan and Sun, Yi and Ma, Shugen and Liu, Handi},
	month = aug,
	year = {2017},
	pages = {1457--1462},
}

@misc{WHO2021Falls,
  author       = {{World Health Organization}},
  title        = {Falls},
  howpublished = {\url{https://www.who.int/news-room/fact-sheets/detail/falls}},
  month        = apr,
  day          = {26},
  year         = {2021},
  note         = {Accessed: 2026-05-21}
}

@techreport{cpwr2018chartbook,
  title        = {The Construction Chart Book: The U.S. Construction Industry and Its Workers},
  author      = {{CPWR}},
  edition      = {6th},
  month        = feb,
  year         = {2018},
  url          = {https://www.cpwr.com/wp-content/uploads/The_6th_Edition_Construction_eChart_Book.pdf},
}

@online{osha_stop_falls,
  author  = {{Occupational Safety and Health Administration}},
  title   = {Fall Prevention Campaign},
  year    = {2025},
  url     = {https://www.osha.gov/stop-falls},
  urldate = {2025-06-12},
}

@article{CHARLES2018125,
title = {Vibration and Ergonomic Exposures Associated With Musculoskeletal Disorders of the Shoulder and Neck},
journal = {Saf. Health Work},
volume = {9},
number = {2},
pages = {125-132},
year = {2018},
issn = {2093-7911},
doi = {https://doi.org/10.1016/j.shaw.2017.10.003},
author = {Luenda E. Charles and Claudia C. Ma and Cecil M. Burchfiel and Renguang G. Dong},
}

@misc{baubot_patent,
  author = {Mieth, Gabriel},
  title = {Mobile robot with configurable adapter},
  howpublished = {European Patent No.\ EP4574354A1},
  year = {2025},
}

@article{augugliaro_flight_2014,
	title = {The {Flight} {Assembled} {Architecture} installation: {Cooperative} construction with flying machines},
	volume = {34},
	issn = {1941-000X},
	shorttitle = {The {Flight} {Assembled} {Architecture} installation},
	doi = {10.1109/MCS.2014.2320359},
	number = {4},
	urldate = {2025-07-24},
    journal = {IEEE Control Syst. Mag.},
    author = {Augugliaro, Frederico and Lupashin, Sergei and Hamer, Michael and Male, Cason and Hehn, Markus and Mueller, Mark W. and Willmann, Jan Sebastian and Gramazio, Fabio and Kohler, Matthias and D'Andrea, Raffaello},
	month = aug,
	year = {2014},
	pages = {46--64},
}

@article{zhang_aerial_2022,
	title = {Aerial additive manufacturing with multiple autonomous robots},
	volume = {609},
	copyright = {2022 The Author(s), under exclusive licence to Springer Nature Limited},
	issn = {1476-4687},
	doi = {10.1038/s41586-022-04988-4},
	language = {en},
	number = {7928},
	urldate = {2025-07-24},
	journal = {Nature},
	author = {Zhang, Ketao and Chermprayong, Pisak and Xiao, Feng and Tzoumanikas, Dimos and Dams, Barrie and Kay, Sebastian and Kocer, Basaran Bahadir and Burns, Alec and Orr, Lachlan and Alhinai, Talib and Choi, Christopher and Darekar, Durgesh Dattatray and Li, Wenbin and Hirschmann, Steven and Soana, Valentina and Ngah, Shamsiah Awang and Grillot, Clément and Sareh, Sina and Choubey, Ashutosh and Margheri, Laura and Pawar, Vijay M. and Ball, Richard J. and Williams, Chris and Shepherd, Paul and Leutenegger, Stefan and Stuart-Smith, Robert and Kovac, Mirko},
	month = sep,
	year = {2022},
	pages = {709--717},
}

@article{hong_agile_2022,
	title = {Agile and versatile climbing on ferromagnetic surfaces with a quadrupedal robot},
	volume = {7},
	doi = {10.1126/scirobotics.add1017},
	number = {73},
	urldate = {2025-07-24},
    journal = {Sci. Robot.},
    author = {Hong, Seungwoo and Um, Yong and Park, Jaejun and Park, Hae-Won},
	month = dec,
	year = {2022},
	pages = {eadd1017},
}

@article{hernando_romerin_2022,
	title = {{ROMERIN}: {A} new concept of a modular autonomous climbing robot},
	volume = {19},
	issn = {1729-8806},
	shorttitle = {{ROMERIN}},
	doi = {10.1177/17298806221123416},
	language = {EN},
	number = {5},
	urldate = {2025-07-27},
    journal = {Int. J. Adv. Robot. Syst.},
    author = {Hernando, Miguel and Gambao, Ernesto and Prados, Carlos and Brito, Daniel and Brunete, Alberto},
	month = sep,
	year = {2022},
	COMMENTEDOUT_pages = {17298806221123416},
}

@misc{hausbots_patent,
  author = {Abry, Zeke and Abraham, Neil and Cornes, Jack},
  title = {Wall climbing device},
  howpublished = {U.K. Patent Application No. GB2637023A},
  year = {2025},
}

@misc{canvas_app_patent,
  author       = {Albert, Kevin B. and Allen, Thomas F. and Bennetsen, Henrik 
                  and Hein, Gabriel F. and Pedersen, Josephine Marie 
                  and Pompa, Jonathan B. and Telleria, Maria J. 
                  and Yan, Charlie and Yoel, Alana G. R. 
                  and Flannery, Miles J. and Tonoyan, Henry},
  title        = {Automated drywall sanding system and method},
  howpublished = {U.S. Patent Application No.\ US20180281143A1},
  date         = {2018-10-04},
}

@article{hack_structural_2020,
	title = {Structural stay-in-place formwork for robotic in situ fabrication of non-standard concrete structures},
	volume = {115},
	issn = {09265805},
	shorttitle = {Structural stay-in-place formwork for robotic in situ fabrication of non-standard concrete structures},
	doi = {10.1016/j.autcon.2020.103197},
	language = {en},
	urldate = {2025-07-27},
	journal = {Autom. Constr.},
	author = {Hack, Norman and Dörfler, Kathrin and Walzer, Alexander Nikolas and Wangler, Timothy and Mata-Falcón, Jaime and Kumar, Nitish and Buchli, Jonas and Kaufmann, Walter and Flatt, Robert J. and Gramazio, Fabio and Kohler, Matthias},
	month = jul,
	year = {2020},
	pages = {103197},
}

@article{alexi_cooperative_2024,
	title = {Cooperative augmented assembly ({CAA}): augmented reality for on-site cooperative robotic fabrication},
	volume = {8},
	issn = {2509-8780},
	shorttitle = {Cooperative augmented assembly ({CAA})},
	doi = {10.1007/s41693-024-00138-6},
	language = {en},
	number = {2},
	urldate = {2025-07-27},
	journal = {Constr. Robot.},
	author = {Alexi, Eleni Vasiliki and Kenny, Joseph Clair and Atanasova, Lidia and Casas, Gonzalo and Dörfler, Kathrin and Mitterberger, Daniela},
	month = oct,
	year = {2024},
	pages = {28},
}

@article{asadi_pictobot_2018,
	title = {Pictobot: {A} {Cooperative} {Painting} {Robot} for {Interior} {Finishing} of {Industrial} {Developments}},
	volume = {25},
	issn = {1558-223X},
	shorttitle = {Pictobot},
	doi = {10.1109/MRA.2018.2816972},
	number = {2},
	urldate = {2025-07-27},
    journal = {IEEE Robot. Autom. Mag.},
    author = {Asadi, Ehsan and Li, Bingbing and Chen, I-Ming},
	month = jun,
	year = {2018},
	pages = {82--94},
}

@article{kindle_enhancing_2025,
	title = {Enhancing {Robotic} {Precision} in {Construction}: {A} {Modular} {Factor} {Graph}-{Based} {Framework} to {Deflection} and {Backlash} {Compensation} {Using} {High}-{Accuracy} {Accelerometers}},
	volume = {10},
	issn = {2377-3766},
	shorttitle = {Enhancing {Robotic} {Precision} in {Construction}},
	doi = {10.1109/LRA.2024.3506276},
	number = {1},
	urldate = {2025-07-27},
	journal = {IEEE Robot. Autom. Lett.},
	author = {Kindle, Julien and Loetscher, Michael and Alessandretti, Andrea and Cadena, Cesar and Hutter, Marco},
	month = jan,
	year = {2025},
	pages = {288--295},
}

@misc{gtsam,
  author       = {Frank Dellaert and Contributors},
  title        = {{GTSAM 4.3a0}},
  year         = 2025,
  month        = May,
  note          = {{Git Repository}, \url{https://github.com/borglab/gtsam} {Accessed}: 2026-05-21}
}

@misc{jaibot_patent,
  author = {Halvorsen, Havard and Henninge, Tom Asle and Fagertun, Konrad},
  title = {Mobile robotic drilling apparatus and method for drilling ceilings and walls},
  howpublished = {U.S. Patent No. US20220001461A1},
  year = {2022},
}

@article{lanegger2023chasing,
  title={Chasing millimeters: design, navigation and state estimation for precise in-flight marking on ceilings},
  author={Lanegger, Christian and Pantic, Michael and B{\"a}hnemann, Rik and Siegwart, Roland and Ott, Lionel},
  journal = {Auton. Robots},
  volume={47},
  number={8},
  pages={1405--1418},
  year={2023},
  publisher={Springer}
}

@inproceedings{dautzenberg_perching_2023,
	address = {Detroit, MI, USA},
	title = {A {Perching} and {Tilting} {Aerial} {Robot} for {Precise} and {Versatile} {Power} {Tool} {Work} on {Vertical} {Walls}},
	isbn = {978-1-66549-190-7},
	doi = {10.1109/IROS55552.2023.10342274},
	language = {en},
	urldate = {2023-12-18},
    booktitle = {Proc. IEEE/RSJ Int. Conf. Intell. Robots Syst. (IROS)},	
    publisher = {IEEE},
	author = {Dautzenberg, Roman and Küster, Timo and Mathis, Timon and Roth, Yann and Steinauer, Curdin and Käppeli, Gabriel and Santen, Julian and Arranhado, Alina and Biffar, Friederike and Kötter, Till and Lanegger, Christian and Allenspach, Mike and Siegwart, Roland and Bähnemann, Rik},
	month = oct,
	year = {2023},
	pages = {1094--1101},
}

@inproceedings{ercan_automated_2019,
	title = {Automated {Localization} of a {Mobile} {Construction} {Robot} with an {External} {Measurement} {Device}},
	doi = {10.22260/ISARC2019/0124},
    booktitle = {Proc. 36th Int. Symp. Automat. Robot. Constr. (ISARC)},
	author = {Ercan, Selen and Meier, Sandro and Gramazio, Fabio and Kohler, Matthias},
	year = {2019},
	pages = {929-936},
}

@article{Kabsch1976,
  author  = {Kabsch, W.},
  title   = {A solution for the best rotation to relate two sets of vectors},
  journal = {Acta Crystallogr. A},
  year    = {1976},
  volume  = {32},
  number  = {5},
  pages   = {922--923},
  doi     = {10.1107/S0567739476001873},
}

\end{document}